# A Compositional Neuro-Controller for Advanced Motor Control Tasks

Kirill Makukhin

*Abstract*—Humans and animals developed a sophisticated motor control apparatus and there is much evidence that it has a modular structure. The modularity offers a range of benefits, e.g. ability to learn dissociable motion styles without interference and forgetting, fast adaptation and de-adaptation to changes conditions. However in robotics, building a controller that can efficiently incrementally learn new motion styles and provide switching between them is a formidable challenge. In this paper we address the problem by proposing a novel biologically inspired compositional neuro-controller. We have shown that the compositional controller is able to reproduce a set of trajectories more efficiently comparing to a simple controller, exploiting incremental learning benefits. Second, we have demonstrated that the proposed controller is able to learn different locomotion styles and switch between them in a simulated robot-snake.

## I. Introduction

HUMANS and animals developed a sophisticated motor control apparatus that is able to learn efficiently dissociable movements. For example we, human, can walk in an energy efficient manner, or we can run achieving maximal possible locomotion speed. Thus, the same brain circuitry is able to provide "switching" between styles, while keeping ability to rapidly adapt to new environments and/or tasks without loosing previously learned skills.

In contrast, the current robots are limited in their ability to interact with the real world, learn and adapt, because their control mechanism is often relied on analytical models of the body and environment with manually fine-tuned parameters. A relatively newly emerged neuro-control domain is dedicated to alleviate this important problem by mimicking in robots the neurological processes that endow animals and humans with nearly optimal performance.

A straight-forward approach to the construction of a neuro-controller capable to generate different locomotion styles, would be to add to a large neural network a special input that switches styles (e.g. [1]). However, such a homogeneous controller seems to be difficult to scale to expand both the degrees of freedom of the robot and the number of styles. Furthermore, the addition of novel styles will "erase" previously learned trajectories unless the training procedure interleaves the new training sent with the old ones.

In contrast, it is believed that the motor abilities of animals are governed by the existence of multiple "stacked" internal controllers that compose their outputs to achieve a reach repertoire of behaviours [2]. The next section provides a brief review of evidence of the existence of multiple models in animals from the neuroscience literature.

Despite of the massive body of research on motor control in neuroscience, the field of robotics/machine learning has much less success in building systems that utilises such compositional controllers. One of the first related work (albeit in the pattern recognition domain) was published in [3], where the system's output was a blend of the outputs of a few separated sub-networks performed by a gating structure. Such a model has limited application because, the decomposition must be known a priori (each sub-network was trained separately).

In a followed work [4], also from the recognition domain, a proposed system contains multiple feed forward networks (experts) and a gating network that decides which experts are appropriate to the current task. Unlike the previous paper, in this system the gating network could learn how to allocate experts to different tasks. In addition, the system restricts the expert parameters update to few of them that provide the most contribution to the current task. Furthermore, the outputs of the experts were normalised by the mean of the softmax function that encourage competition between the experts (localisation). It was shown that such a system has a few significant advantages over homogeneous networks. In particular, it drastically reduces the interference between tasks (i.e. forgetting previous knowledge), while improves generalisation within similar tasks.

In the motor control domain, Doya et al. [5] proposed a multiple model-based reinforcement learning framework that performs decomposition of a complex task. Each separate module contains two structures: a reinforcement learning controller and a predictor that estimates the "responsibility" of the module and is used to weight its output in the final control signal. The researchers showed that such decomposition can speed up learning of tasks that are nonlinear or nonstationary in space and time. However, there is no evidence that the method is scalable to real world applications, such as robot locomotion or flight control. Furthermore, the approach implicitly assumes that the predictor and RL policy are learned synchronously. That is, the responsibility signal that is based

K. Makukhin, was with the University of Queensland, Saint Lucia, QLD 4072, Australia, (e-mail: k.makukhin@webage.net.au). However, this work was done at author's own time and using his own resources.

on the predictor's error, matches the efficacy of the policy. Intuitively, it is not always the case because the two systems are weakly coupled. Finally, this framework did not address the problem of switch between tasks, which is the major research question of the current paper.

An extension of the decompositional framework to imitation learning was proposed in [6]. The researchers have introduced a system called the mixture of motor primitives (MoMP). In this framework the motor control signal is generated as a weighted sum of primitive action policies. The policies are recorded by kinaesthetic demonstration, and the gating parameters are learned with a variation of the reinforcement learning algorithm. Thus, the system decomposes complex motor task onto the primitive actions. The researchers have shown spectacular results, e.g. Ping-Pong game playing robot. However, the system relies on the quality and completeness of the primitives introduced by teacher. Furthermore, it is an open question if this approach is scalable to highly dissociable actions as long as the homogenous gating network is the only trainable part and the primitives is not adjustable.

A different approach was taken by Cully & Mouret [7]. Their hexapod robot has learned a range of simple behaviours, such as walking forward with different degree of turning. The controller comprised a combination of oscillators, parameterised by an amplitude and phase shift, and trained by a multi-objective optimisation genetic algorithm. It is important to note that the system does not perform mixing of the oscillators (i.e. sub-controllers). That is, every new behaviour relies on a separate sub-controller and there is no generalisation across the learned behaviours, which perhaps will lead to the lack of scalability for real world applications.

Doncieux & Meyer [8] showed that the controller's layout could be learned by experience as well. They proposed a framework "ModNet" for the evolving of a modular networks for control tasks (e.g. cart-pole balancing problem). ModNet discovered useful modules that were propagating through generations and were being reused over new trials, but it is unclear if this approach is scalable to the real world environment. While the idea of evolving the layout is appealing, it worth to note that the animal's brain was/is exposed to at least two optimisation processes: the evolution that has formed the modularity of the brain over millions years, and life-time learning that allows to optimise "weights" within the already formed structure. Thus, the feasibility of combining these two processes into one is the matter of further studies.

The rest of the paper is organised as following. First, we briefly review the evidence for multiple internal modules in the brain. Then we introduce the compositional controller and describe two experiments illustrating the proposed model specific features. Finally, we discuss the advantages and limitations of the compositional controller, comparing to other approaches.

## II. EVIDENCE FOR MULTIPLE INTERNAL MODELS

Strong experimental evidence from neurophysiological studies suggest that instead of keeping one huge model of all behaviours (or tools, or contexts), organisms learn multiple internal models for every distinguishable behaviour and combine or switch them depending on conditions. For example, Ghahramani & Wolpert [9] studied eye movement control in human subjects. They concluded that faster de-adaptation to a normal condition and re-adaptation to a previously learned condition could only be achieved by retaining multiple models. Furthermore, they suggested that the brain may employ task decomposition, as opposite to simple switching, as well as that learning of models may be facilitated by combining stored modules.

In neuroscience, researchers suggested [10] that the localised activity of small spots in the cerebellum during the use of different tools could represent multiple models for different objects and environmental settings. Another study on frogs and rats [11] has demonstrated that the premotor circuits within spinal cord seems to have a modular structure where the modules form a "vocabulary" that could be added to each other to provide a rich set of behaviours.

Flanagan et al. [12] found clear support for the composition hypothesis suggesting the brain can effectively combine multiple pre-learned internal sensorimotor models for efficient handling a novel task. The researchers also found partial support for the reverse process: decomposition of a complex task into separate appropriate internal models.

It is still an open question on what parts of the brain accommodate multiple models. Recent research suggests that the cerebellum is very promising candidate for internal dynamics and kinematics models, while cerebellar and motor/premotor cortices might take part in task decomposition and/or action composition, e.g. [12]–[16]. In addition, there is evidence that the spinal cord can accommodate motor primitives and combine them into more complex actions, [17], [11].

## III. COMPOSITIONAL CONTROLLER

This paper proposes a compositional controller that is able to incrementally learn a repertoire of motor behaviours without catastrophic or gradual forgetting previously learned skills.

The compositional controller consists of several sub-controllers that receive inputs relevant to the current system state and provide their output contributions to the system's motor control action. These action contributions are combined into the output by the mean of a multiplicative gating network with a linear transfer function, Fig. 1.

Mathematically speaking, the system's output could be described as following:

$$out = \sum_{s=1}^{N} w_{ks} o_s, \qquad (1)$$

where $o_s$ is the sub-controllers' output (action contribution) and $k$ is the current task index. The gating weights table $w_{ks}$ has the number of rows equal to the number of tasks, and each row is a weight vector with $N$ entries, one for each sub-controller.

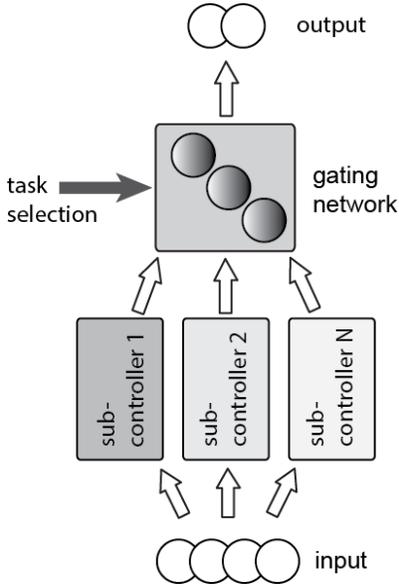

Fig. 1. The compositional controller layout. The sub-controllers receive the current state signal and generate their action contributions that later are combined into output with the gating network. The gating network has separate weight sets for each task.

The sub-controllers are fully-connected recurrent neural networks (RNN). RNNs are known to be able, in principle, to approximate arbitrary dynamical system [18]. In our implementation, each neuron receives all input signals and some neurons are considered as output nodes, Fig. 2. The neurons have tanh non-linearity function (2).

$$o_i^t = tanh\left(\sum_j^{all\ neurons} v_{i,j} o_j^{t-1} + \sum_k^{all\ inputs} u_{i,k} x_k^t\right) \quad (2)$$

The sub-controllers are dedicated and being trained each at its own task, but previously trained controllers may contribute to the output as well through the gating network. The learning procedure sequentially chooses a new task and adapts the task-specific sub-controller's parameters along with the all weights of the gating network relevant to the task[1]. As a result, the algorithm yields an array of trained RNNs and a table of gating weights.

The weights optimisation is performed with the RBM-ES algorithm [19]. This algorithm belongs to the evolution strategies family and yields good results on neural network evolving tasks due to its wide adaptability and multi-modal nature. Briefly, the RBM-ES generates new candidate solutions from a meta-model that is maintained by exploiting the search history. The meta-model is formed by a restricted Boltzmann machine with Gaussian visible units [20]. Perhaps, other black-box optimisation algorithms could be used instead, but this question is out of the paper scope.

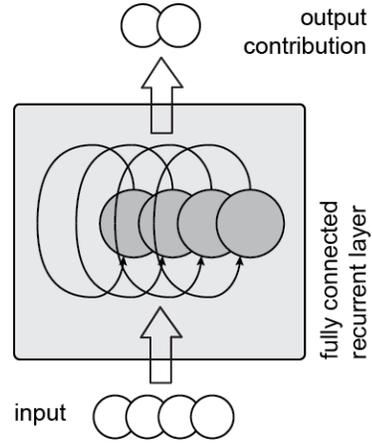

Fig. 2. The sub-controller consists of a fully connected recurrent network. The input signals arrive to each neuron, and some of the neurons project their outputs to the sub-controller output.

The genotype for the RBM-ES algorithm (the sequence of weights being trained) is formed by concatenating both current RNN's parameters and the vector of gating coefficients g specific for the current task. The gating coefficients before applying to the gating network are normalised with the softmax function:

$$w_{ks} = \begin{cases} \dfrac{e^{\tau g_s}}{\sum_N e^{\tau g_s}}, & if\ s \leq k \\ 0, & otherwise \end{cases} \quad (3)$$

where $w_{ks}$ is the final gating weights and $g_s$ is the parameters suggested by the optimiser.

It is important to notice that at the time when the $s^{th}$ sub-controller is being trained, the weights of mixing yet un-trained RNNs must be set to zero, as shown in equation (3). Otherwise, these RNN's contributions will affect the performance of the prior tasks when they will have been trained.

The softmax pre-processing of the coefficients enforce a competition between controllers. That is, the system supposed to re-use existing controllers as much as they could contribute. The inverse temperature $\tau$ governs the sharpness of the competition, i.e. the extent of involvement of previously learned controllers.

## IV. INCREMENTAL COMPOSITIONAL LEARNING OF SIMPLE TASKS

### A. Experiment design

In this experiment, we were exploring the process of compositional incremental learning on a range of relatively simple tasks: learning to reproduce periodical functions given a cosine input signal.

Specifically, the time required for accomplishing the learning of target functions was compared for two scenarios: when a single controller is trained from scratch and when the training is applied to a sub-controller of a compositional

---

[1] except for the weights corresponding to un-trained yet sub-controllers, which are forced to be zero.

system that had been pre-trained with several periodical basis[2] functions.

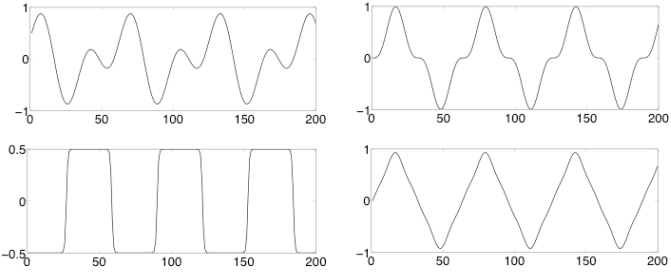

Fig. 3. The visualisation of the four target functions used in the first experiment.

Table 1 lists the target and basis functions chosen for the experiment (see also Fig. 3).

TABLE 1
TARGET FUNCTIONS USED IN THE FIRST EXPERIMENT

| Function | Equation |
|---|---|
| Basis function B1 | $y = \sin x$ |
| Basis function B2 | $y = \sin 2x$ |
| Basis function B3 | $y = \sin 3x$ |
| Target function T1 | $y = \frac{\cos x}{2} + \frac{\sin 2x}{2}$ |
| Target function T2 | $y = \sin x^3$ |
| Target function T3 | $y = \text{sigm}\left[\sin\left(x + \frac{\pi}{6}\right) * 20\right] - 0.5$ |
| Target function T4 | $y = \frac{8}{\pi^2}\left[\sin x - \frac{1}{9}\sin 3x + \frac{1}{25}\sin 5x\right]$ |

In this setup, the sub-controller' RNN consist of seven neurons, one input and one input. Thus, each RNN has seven input weights $u_{i,1}$, which is 7 by 7 recurrent weights $v_{ij}$, 56 in total. The number of the neurons within the RNN was chosen arbitrary, keeping in mind that the RNN need to be powerful enough to quickly learn target functions, but not too large to reduce computations.

To engage oscillating behaviour of RNNs and simplify training we applied a cosine signal to the input. Albeit it is not strictly required, it speedups training and, importantly, it provides a meaningful input to the system, which is desirable to be consistent with our following experiments.

The genotype was formed by concatenating the set of all 56 RNN's parameters and the vector of gating coefficients $g$. We found that using quite high value of the inverse temperature $\tau$ engages more competition and improves the convergence speed: we set it to 4.0.

The optimisation was performed with the RBM-ES algorithm, minimising the squared error between the target output and the output of the compositional controller. The training is continued until the error drops down to 0.9 or the number of epochs reaches 20000.

---

[2] We call these functions "basis" functions, but strictly speaking they could be arbitrary and do not require any specific properties such as being "orthogonal".

## B. Results

The goal of the experiment was to explore the process of compositional learning on a relatively simple testbed. The experiment comprised two stages; at the both stages we trained the controller to reproduce periodical functions, minimising the squared error loss function.

For the first stage we had selected simple functions that seem to be helpful to form a "basis" for the learning of following more complicated target functions (see Table 1). This helps to emphasize the cooperative contribution of previously learned sub-controllers in the solving a novel problem, demonstrating incremental learning.

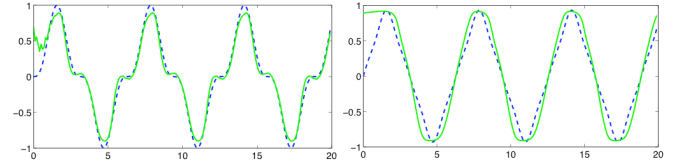

Fig. 4. Examples of learned (solid line) vs. target (dashed line) functions. Left plot: T2, right plot: T4. Best viewed in colour.

An example of learned vs. target functions is shown in Fig. 4, and the performance comparison averaged over 25 trials is presented in Fig. 5. The first bars in the groups (blue) show the number of epochs required to learn target functions when the compositional controller has only one sub-controller, trained for this particular target function. The second bars in the groups (green) represent the training time for the case when the compositional controller has been pre-trained on the three "basis" functions (Table 1). As it can be seen on the chart, pre-training the controller results in faster learning for all target functions.

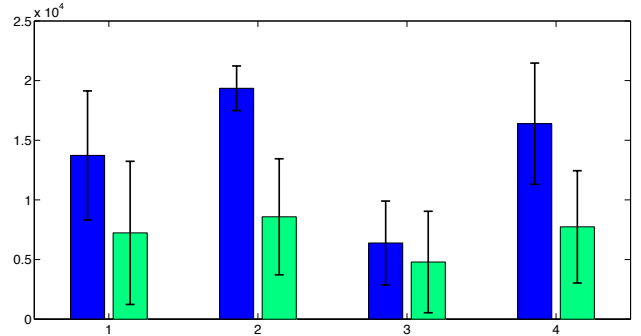

Fig. 5. The number of epochs required to learn four target functions from scratch (the first bars in the groups) and after pre-training the compositional controller with "basis" functions (the second bars). Averaged over 25 trials. Best viewed in colour.

Fig. 5 illustrates how the system re-uses previously learned experience. It shows the final gating network weights after the training in log scale. Each group corresponds to a separate task, i.e. it represents a vector of sub-controllers' mixing coefficients that were trained (along with the task-dedicated sub-controller) to reproduce its target function. The bars B1-B3 show the mixing amount of the pre-trained "basis" functions, and the bars T1-T4 correspond to the mixing proportion of the target functions. For example, Task 2 actively uses the sub-controller that was trained on B2 and only slightly uses the B2 sub-controller.

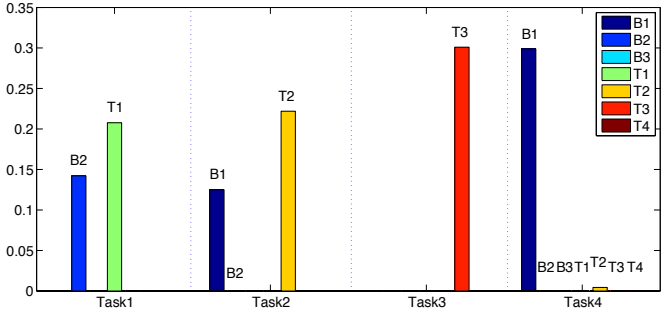

Fig. 6. The gating network weights after training all target functions. Each of the four groups represents weights learned to reproduce target functions (T1-T4) in log scale. The bars correspond to the proportion of mixing the outputs of sub-controllers that have previously learned the "basis" (B1-B3) and preceding target functions (T1-T4). The bars that have nearly zero value are not labelled. Best viewed in colour.

The first target function T1 substantially re-uses the "basis" B1 in addition to its dedicated controller T1. Similar picture is for the target T2 that mostly uses B1 and little bit of B2. In contrast, the target T3 has evolved it's own sub-controller only. It is interesting to note that T4 task is almost fully decomposed to the preceding sub-controllers and allocate literally no share to its own controller T4.

## V. LEARNING DISSOCIABLE LOCOMOTION STYLES FOR A ROBOT-SNAKE

### A. Experiment design

In this experiment we applied the proposed system to the problem of locomotion control for a simulated robot-snake. Not only was the goal of the experiment was to achieve efficient locomotion, but also to create a system that provides with a *selectable* motor skill repertoire in the close-to-real-world environment.

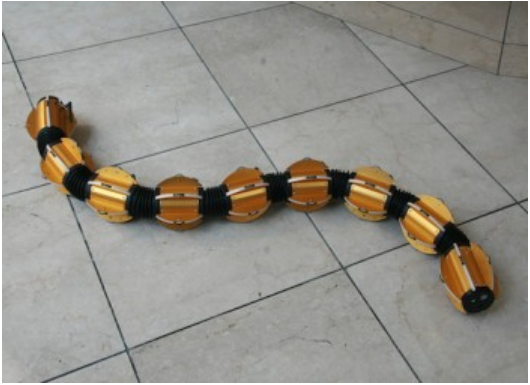

Fig.7. A photo of the ACM-R5H robot-snake. With permission of Hibot Corp.

The simulation was performed at the V-Rep simulator [21] with the model of a real robot-snake ACM-R5H designed by Hibot Corporation, Japan. The robot consists of nine lined-up segments connected to each other with active joints. Thus, the robot has eight control servos. Every segment of the robot also has a few sets of passive wheels to ensure that the ground friction forces acting on the robot are anisotropic in the tangential (along the locomotion) and normal directions, which is desirable during motion across flat surfaces [22].

The robot is controlled by a central pattern generator (CPG), similarly to the model proposed in [23]. The CPG effectively is a chain of mutually coupled oscillators, one oscillator for each joint. The oscillators are described by a set of differential equations[3]:

$$\dot{\theta} = 2\pi v_i + \sum_j r_j w_{ij} \sin(\theta_j - \theta_i - \varphi_{ij})$$
$$\ddot{r}_i = \alpha_i \left(\frac{\alpha_i}{4}(R_i - r_i) - \dot{r}_i\right)$$
$$x_i = X_i + r_i \cos(\theta_i),$$

where for *i*-th oscillator:
$\theta_i$ - current phase;
$r_i$ - current amplitude;
$v_i$ - intrinsic frequency;
$R_i$ - the attractor amplitude;
$\alpha_i$ - a positive constant;
$X_i$ - an offset from the central position.

The coupling between neighbour oscillators is defined by the weight $w_{ij}$ and phase bias $\varphi_{ij}$; the value of $x_i$ represents the output of the oscillator. In our experiment we set $v_i = 1$Hz for all oscillators, and the bias values $\varphi_{ij}$ are equal to $2\pi/N$ for head-to-tail connections and $-2\pi/N$ for the opposite direction, where $N$ is the number of body segments. The weights $w_{ij} = 20$ for all connections. The time step in the simulation is 50 ms.

The compositional controller has similar layout to the controller in the previous experiment (Fig. 1), but it receives no input. The sub-controllers are the tables of the oscillators amplitudes $R_i$ and offsets $X_i$. The optimisation algorithm RBM-ES was applied to the sub-controllers' table values and gating network weights.

The amplitude $R_i$ need to be always positive, while the optimisation algorithm produces both positive and negative values. To handle this, we pass the values that encode amplitudes through logistic function.

The system was trained to achieve three different locomotion styles: moving straight, turning left and turning right. The optimiser objective function (reward) was evaluated for a whole episode:

$$R = R_{dist}\log(d + 1) + \beta R_{side}s.$$

The first term is a distance component, where $d$ is the distance travelled by the robot in metres; we found that the logarithmic scale encourages the robot to find efficient locomotion more quickly. Snake-type robots cannot turn to a side on-the-spot and require some turning radius. This is why there is a second sidewise distance component that is proportional to the side shift coordinate $s$. $\beta$ is a sign correction parameter:

---

[3] For more detailed description of this type of CPG based on coupled oscillators we refer the reader to [23] and [27].

$$\beta = \begin{cases} -1, & turning\ right \\ +1, & turning\ left \\ 0, & going\ straight \end{cases}$$

The following rewards were used in the experiment: $R_{dist}$=10, $R_{side}$=20.

Each episode ends after $E$=120 time steps. When the goal is turning to a side, then the episode can also end after the robot head has achieved the angle of 120 degrees to the direction of turning.

*B. Results*

The example of resulting trajectories for moving forward, turning left and turning right are presented in Fig. 8. The compositional controller has learned three dissociable locomotion styles that could be selected independently or, possibly, transitioned from one to another. We encourage the reader to watch the video of the experiment results [24].

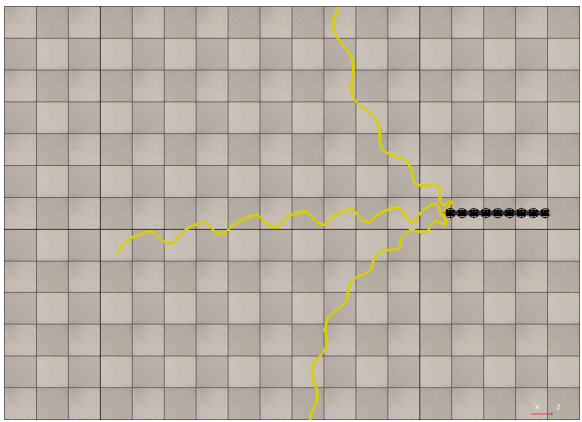

Fig. 8. Example trajectories of the robot's head for three dissociable styles: going straight, turning left and turning right.

Decomposition of the tasks in this experiment can be seen in Fig. 9. Each bar group represents a different task – forward locomotion, turning left and turning right. Each task readily re-uses existing controller when possible[4]. For example, the third task adjusts parameters of the 3$^{rd}$ sub-controller, but also re-uses both the 1$^{st}$ and 2$^{nd}$ sub-controllers.

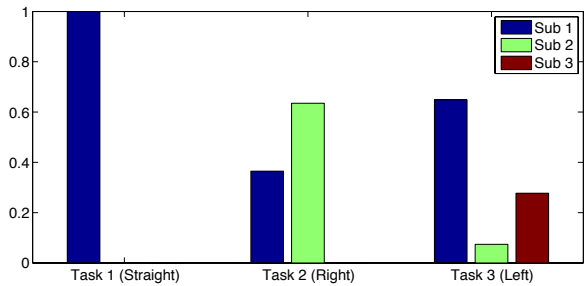

Fig. 9. An example of the tasks' decomposition. Each bar group represents different tasks. The bars within group shows the extend of usage of all three sub-controllers.

It is worth noticing, that the dimensionality of the robot's

---

[4] Recall that the sub-controllers are trained in sequence. Thus, for example, when the controller is trained on the second task, it only can re-use the 1$^{st}$ sub-controller, but not the 3$^{rd}$ one that is undefined yet.

actuation is relatively low and the training required only 600 simulation episodes for each task.

## VI. DISCUSSION

While substantial body of research is focused on learning a single-action controller, e.g. locomotion on a straight line or keeping stable gait regimes, the focus of this paper is on a more general controller that would allows switching or transitioning between different motion styles (e.g. walk, trot, canter and gallop, or curving opposite directions), which seems to be a more complicated task.

The point of departure in the proposed model was the architecture suggested by Doya [5], where a controller with multiple internal modules and mixed outputs was designed. The contribution of each module before mixing was weighted by "responsibility" signal. The controller in Doya's experiment was *decomposing* tasks into more or less "orthogonal" primitive actions. The researchers showed, for example, that the inverted pendulum problem could be solved more efficiently with the suggested decomposition framework rather than in plain reinforcement learning.

In contrast, our approach allows every novel task to initiate learning of a separate sub-controller, while keeping re-using of already learned ones (when they could fit at any extend to the current task). That is, strictly speaking, no decomposition is *enforced*, albeit it usually happens at some degree as shown in Fig. 6 and Fig. 9. This could be also viewed as if the sub-controllers incrementally learn overcomplete basis of non-orthogonal simple actions[5]. While our approach might seem to be less efficient way of storing learned actions, it has a significant advantage over the Doya's model in scalability. That is, given that the number of sub-controllers is not limited[6], our system could learn nearly infinite repertoire of actions without forgetting anything[7].

Another model that has some similarities is the MoMP framework, suggested in [6]. It relies on sub-controllers that effectively are pre-recorded by kinaesthetic demonstration primitive actions. In contrast, in our system all simple actions are learned by optimal control. Furthermore, our framework allows incremental learning. That is, new sub-controllers could be added at any time and they can re-use previously learned experience, thus reducing required learning time and resources.

In principle, our framework could incorporate heterogeneous sub-controllers and/or different training methods. Thus, some of the sub-controllers could be, for example, pre-recorded by a demonstrator (similarly to the MoMP) or even hard-wired, providing the compositional controller with a useful basis of simple actions to leverage learning more complicated motor skills.

---

[5] We contrast the actions learned by our sub-controllers with Doya's modules by using the word "simple" instead of "primitive". That is, our simple actions could be quite complicated and not necessarily useful for other actions.

[6] Thanks to the incremental nature of compositional controller training, the complexity of every following sub-controller might become lower comparing to the preceding ones (e.g. in first experiment, the target T4 does not use its corresponding 4$^{th}$ sub-controller at all – see Fig. 6). Thus, even given limited resources, the number of controllers could approach substantially large quantity.

[7] It is worth noticing that another potential problem could be overfitting.

Furthermore, we would speculate that the proposed modular approach should allow to easily exclude rarely used sub-controllers from the system, thus freeing up unused resources. This could be seen as *controlled* pruning the system, where the process only touches specific pre-defined resources, but do not cause a degradation of the quality of useful information.

Our framework learns a separate sub-controller and gating weights for every novel motion style and/or environment context. However, this might pose a question: when it is feasible to consider a task as a novel one, or what is the granularity of the internal models? Wolpert & Ghahramani [25] provided an example where motor commands for lifting an *empty* or *full* milk cartons are governed by two separate controllers. Although such fine division might appear as "wasting" of the brain computational resources at first sight, in practice it helps to successfully deal with two important issues. First, it eliminates the problem of catastrophic or gradual forgetting, a phenomenon peculiar to most connectionist networks [26]. That is, in the case of a compositional controller, novel knowledge never cause degradation of previously learned skills. Second, it provides with a mechanism for fast adaptation and de-adaptation to novel conditions such as changing external force fields affecting limbs movements [2]. Having two separate sub-controllers for even similar contexts allows fast switching between two motor skills instead of re-learning.

The compositional controller provides a mechanism for the execution of dissociable motor consequences. However, obviously there is a need for a higher level meta-system that decompose global tasks (such as how to arrive from point A to point B) to the specific sub-tasks governed by the controller, e.g. walk forward, then turn left, then stop. This important question is out of the scope of the current research, and we leave it for future work.

## VII. Conclusion

In this paper, we proposed a novel biologically-inspired approach that addresses the problem of complex motor control. Specifically, we introduced a compositional neuro-controller that is able to incrementally learn dissociable motion styles and provide switching between them. This problem is know to be more complicated than learning a single-style motion, such as walking in a straight line [7].

The proposed controller has three major benefits. First, learning novel motion trajectories do not cause a degradation of previously learned skills. Such degradation is known to be a common problem in homogenous neural networks, and perhaps, in some multiple-model controllers that do not specifically address this problem.

Second, the compositional controller allows incremental learning with possible decreasing complexity of additional sub-controllers. This allows efficiently re-use gained experience and also might allow mixing different training approaches: e.g. optimal control could be mixed with learning from demonstration and hard-wiring.

Finally, it provides with a mechanism for fast adaptation/de-adaptation to novel conditions, an important feature commonly observed in animals [2], [12].